\documentclass[letterpaper, 10 pt, conference]{ieeeconf}  % Comment this line out if you need a4paper

\IEEEoverridecommandlockouts                              % This command is only needed if 
\usepackage{graphicx} % for pdf, bitmapped graphics files
\usepackage{graphics}
\usepackage{epsfig} % for postscript graphics files
\usepackage{times} % assumes new font selection scheme installed
\usepackage{amsmath} % assumes amsmath package installed
\usepackage{subcaption}
\usepackage{amssymb}  % assumes amsmath package installed
\usepackage{gensymb}
\usepackage{multirow}
\usepackage{makecell}
\usepackage{booktabs,chemformula}
\usepackage{array}
\usepackage{url}
\newcolumntype{C}[1]{>{\centering\arraybackslash}p{#1}}
\newcommand{\tabincell}[2]{\begin{tabular}{@{}#1@{}}#2\end{tabular}}
\newcommand{\etal}{\textit{et al}. }

\newcommand{\darkgrayed}[1]{\textcolor{darkgray}{#1}}
\makeatletter
\newcommand*\titleheader[1]{\gdef\@titleheader{#1}}
\AtBeginDocument{%
  \let\st@red@title\@title
  \def\@title{%
    \vskip-3em
    \bgroup\normalfont\large\centering\@titleheader\par\egroup
    \vskip1.5em\st@red@title}
}
\makeatother

\titleheader{\darkgrayed{This paper has been accepted for publication at the \\ IEEE International Conference on Robotics and Automation (ICRA), Xi’an, 2021.
\copyright IEEE}}

\author{Davide Falanga$^*$, Philipp Foehn$^*$, Peng Lu, and Davide Scaramuzza 
\thanks{$^*$ These authors contributed equally to this manuscript.}
\thanks{This research was supported by the National Centre of Competence in Research (NCCR) Robotics, the SNSF-ERC Starting Grant, the DARPA FLA program. The authors are with the Robotics and Perception Group, Dep. of Informatics, University of Zurich, and Dep. of Neuroinformatics, University of Zurich and ETH Zurich, Switzerland---\url{http://rpg.ifi.uzh.ch}.}%
}

\title{\LARGE \bf
Geometry-aware Compensation Scheme for Morphing Drones
}

\author{Amedeo Fabris, Kevin Kleber, Davide Falanga and Davide Scaramuzza% <-this % stops a space
\thanks{All the authors are with the Robotics and
Perception Group, Dep. of Informatics, University of Zurich, and Dep. of Neuroinformatics, University of Zurich and ETH Zurich, 8050 Zurich, Switzerland
(\protect\url{http://rpg.ifi.uzh.ch}).
This work was supported by the National Centre of Competence in Research (NCCR) Robotics through the Swiss National Science Foundation (SNSF) and the European Union’s Horizon 2020 Research and Innovation Programme under grant agreement No. 871479 (AERIAL-CORE) and the European Research Council (ERC) under grant agreement No. 864042 (AGILEFLIGHT).}%
}

\begin{document}

\maketitle
\thispagestyle{empty}
\pagestyle{empty}

%%%%%%%%%%%%%%%%%%%%%%%%%%%%%%%%%%%%%%%%%%%%%%%%%%%%%%%%%%%%%%%%%%%%%%%%%%%%%%%%
\begin{abstract}
Recent studies have shown that enabling drones to change their morphology in flight can significantly increase their versatility in different tasks. 
%Morphing %multirotors, such as the Foldable Drone, 
%can increase the versatility of drones employing in-flight-adaptive-morphology.
%To further increase precision in their tasks, recent works have investigated stable flight in asymmetric morphologies mainly leveraging the low-level controller. 
%However, the aerodynamic effects embedded in multirotors are only analyzed in fixed shape aerial vehicles and are completely ignored in morphing drones.
%In this paper we investigate and counteract the aerodynamic effects which have a major influence on the Foldable Drone~\cite{falanga2018foldable}, a drone capable of rotating its four arms independently around the central body while flying.
%The investigation is conducted by means of dedicated experiments to analyze the effects that the most compact configurations of the Foldable Drone have on the system, that is, the partial overlap between a propeller and an occlusion underneath it.
In this paper, we investigate the aerodynamic effects caused by the partial overlap between the propellers and the main body of a morphing quadrotor during flight.
We experimentally characterize such effects and design a morphology-aware control scheme to compensate them. % to account for them.
%To guarantee the right trade-off between efficiency and compactness of the vehicle, we propose a simple geometry-aware compensation scheme 
%based on the results of these experiments. 
We demonstrate the effectiveness of our approach by deploying the compensation scheme on a quadrotor that can fold its arms around the main body, comparing it against the same controller without the compensation scheme.
Experimental results show that our compensation scheme can address the loss of thrust due to the overlap between the main body and the propellers, guaranteeing higher tracking accuracy, without requiring complex and computationally expensive aerodynamical models.
%The same set of experiments are performed and compared against one another with and without the compensation scheme offline or during the flight.
To the best of our knowledge, this is the first work counteracting the aerodynamic effects of a morphing quadrotor during flight and showing the effects of partial overlap between a propeller and the central body of the drone.

\end{abstract}

\section*{Supplementary Material}
\noindent
All the videos of the experiments are available at: \\
\url{https://youtu.be/Na0yrkCsC-M}

%%%%%%%%%%%%%%%%%%%%%%%%%%%%%%%%%%%%%%%%%%%%%%%%%%%%%%%%%%%%%%%%%%%%%%%%%%%%%%%%
\section{Introduction}\label{sec:introduction}
In recent years, adaptive morphology~\cite{falanga2018foldable, riviere2018agile, bai2019evaluation, zhao2017whole, zhao2018design, bucki2019design} has increased the already ubiquitous use of multirotors ~\cite{otto2018optimization}, leading to an unprecedented versatility in applications, such as delivery, proximity inspections, and search-and-rescue operations~\cite{delmerico2019current, shakhatreh2019unmanned}. 
The higher maneuverability of morphing drones compared to standard fixed-shape drones in challenging tasks, such as traversing narrow gaps, is achieved without miniaturizing the drone with consequent benefits in terms of flight time, payload, and robustness to air perturbations.
\begin{figure}[!ht]
     \centering
     \begin{subfigure}[t]{0.2\textwidth}
         \centering
         \includegraphics[width=\textwidth]{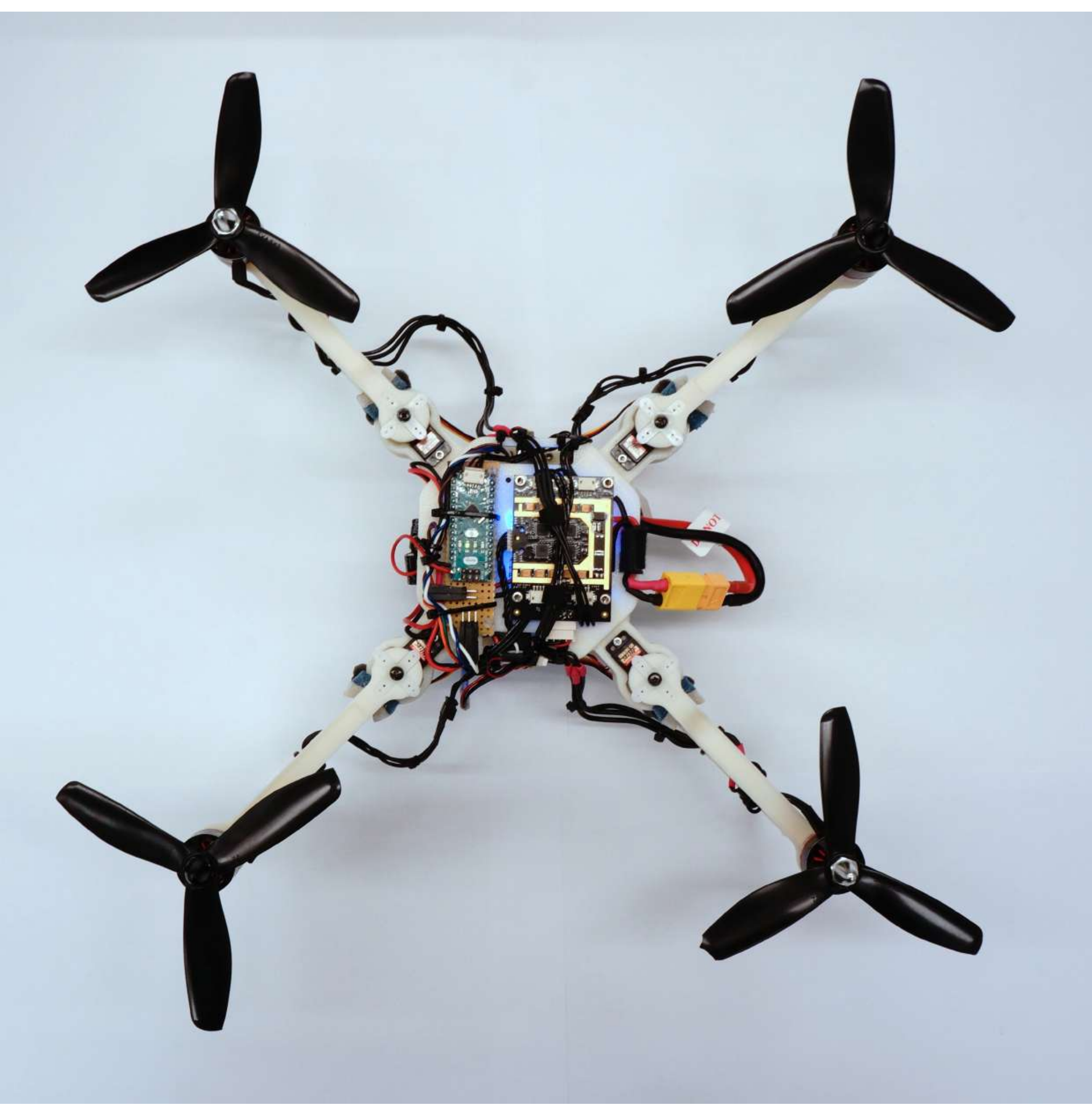}
         \caption{X configuration: no overlap.}
         \label{fig: X_config_table}
         \vspace{2mm}
     \end{subfigure}
     \hspace{-1mm}
     \begin{subfigure}[t]{0.2\textwidth}
         \centering
         \includegraphics[width=\textwidth]{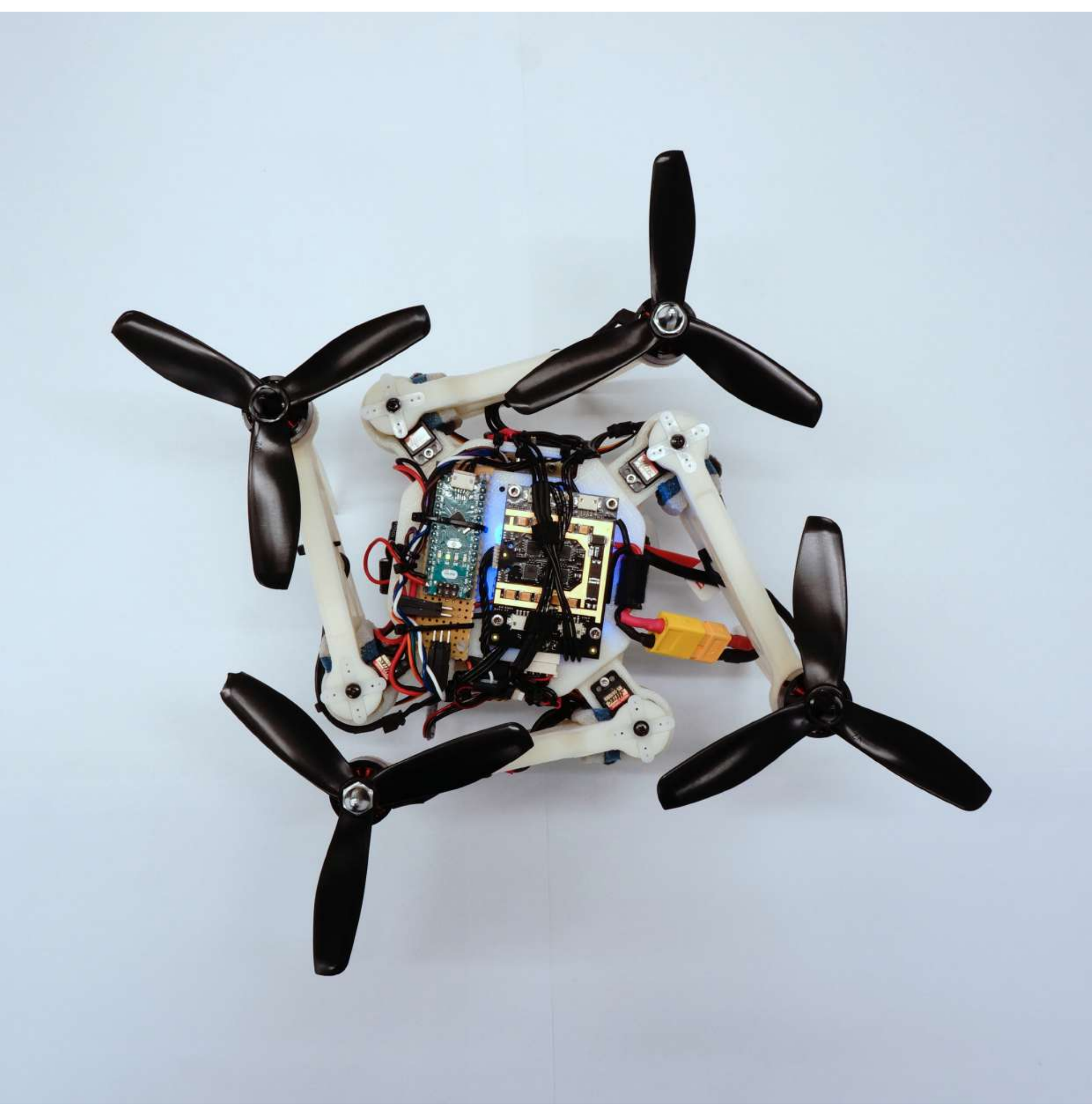}
         \caption{O configuration: overlap.}
         \label{fig: O_config_table}
     \end{subfigure}
     \hspace{5mm}
     \begin{subfigure}[t]{0.2\textwidth}
         \centering
         \includegraphics[width=\textwidth]{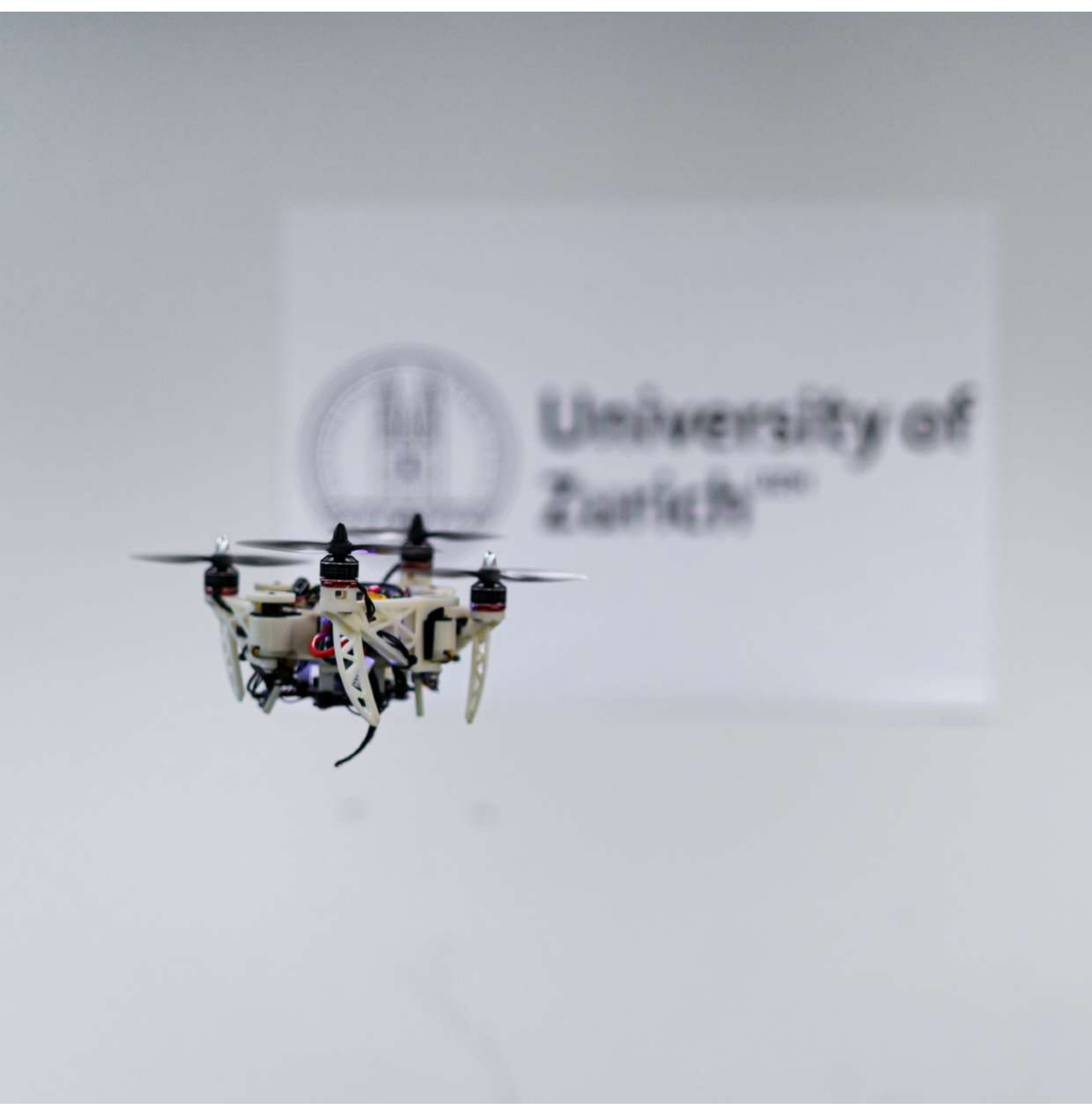}
         \caption{Hovering in O configuration no compensation.}
         \label{fig: X_to_O_nocomp}
     \end{subfigure}
    \begin{subfigure}[t]{0.2\textwidth}
         \centering
         \includegraphics[width=\textwidth]{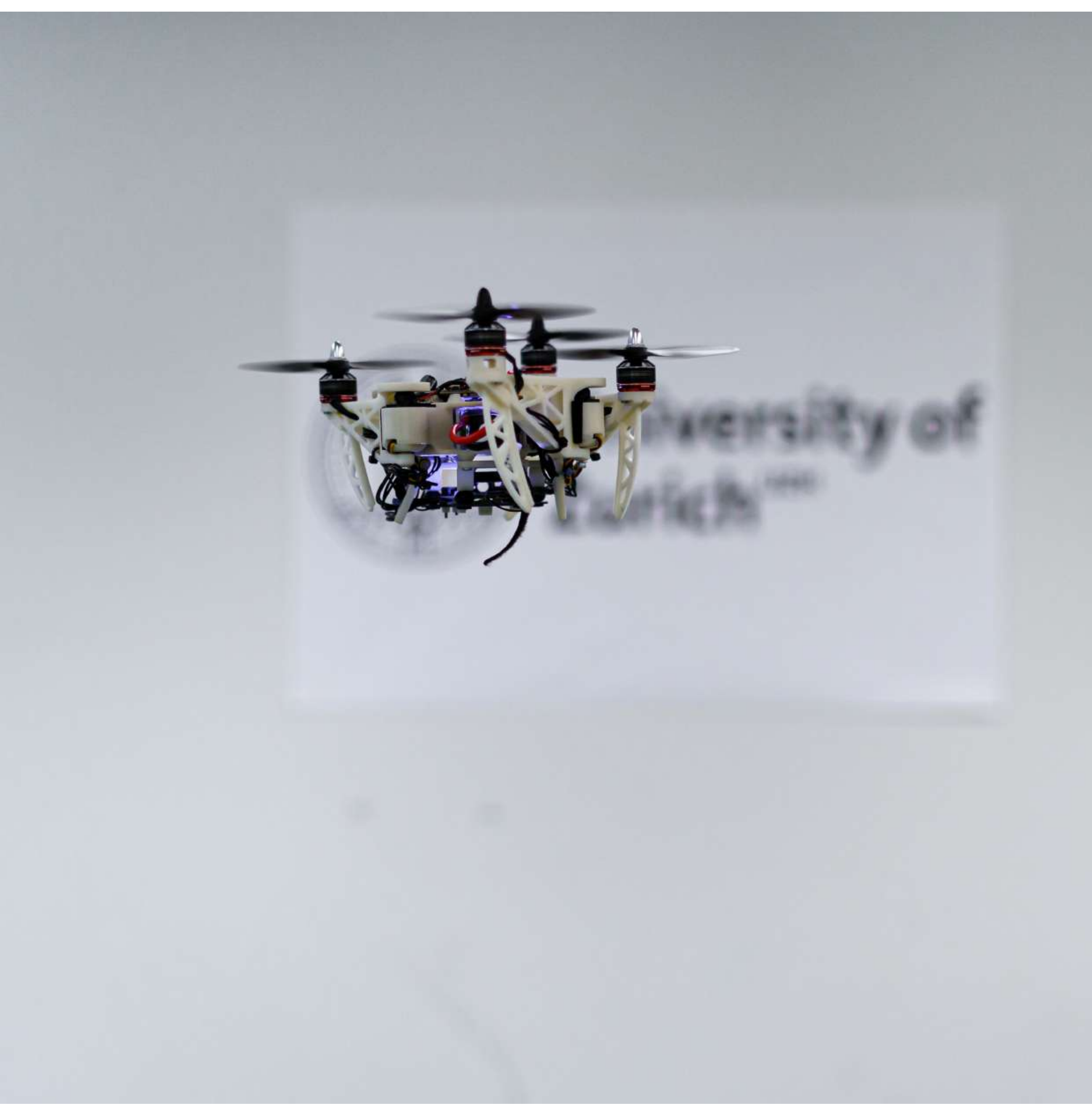}
         \caption{Hovering in O configuration with compensation.}
         \label{fig: X_to_O_comp}
     \end{subfigure}
    \caption{Images of the Foldable Drone~\cite{falanga2018foldable}. (a): the canonical configuration of the Foldable Drone i.e. the X configuration. (b): the most compact configuration of the morphing platform i.e. the O configuration. Close-up views: experiments showing the transition in hovering from X to O configuration with compensation (d) and without compensation (c).}
    \label{fig: hovers}
    \vspace{-3mm}
\end{figure}
The main drawback of morphing is its impact on the aerodynamic properties of these vehicles, as varying geometries can introduce interactions between different parts of the robot.
For example, in the case of a quadrotor that can change its shape, the airflow generated by the propellers can interact with the central body, depending on the morphology assumed by the aircraft.
In the standard X-shape, no overlap exists and, consequently, no interaction happens.
On the contrary, if the robot changes its shape (cf. Fig.~\ref{fig: hovers}), the propellers can overlap with the central body, causing a loss of thrust resulting in a deviation from the target state.
Modeling and compensating for these effects has significant importance in the field of morphing drones since it can dramatically improve their closed-loop tracking capabilities.
However, coping with these effects often requires complex aerodynamical models, which cannot be treated in real-time on small-scale, limited-power computational units that micro aerial vehicles are usually equipped with.
Additionally, these models often require expensive equipment, such as wind tunnels, and powerful machines to elaborate the data.

\subsection{Contributions}
\begin{figure}
\centering

\tabskip=0pt
\valign{#\cr
  \hbox{%
    \begin{subfigure}[b]{.16\textwidth}
    \centering
    \includegraphics[width=\textwidth]{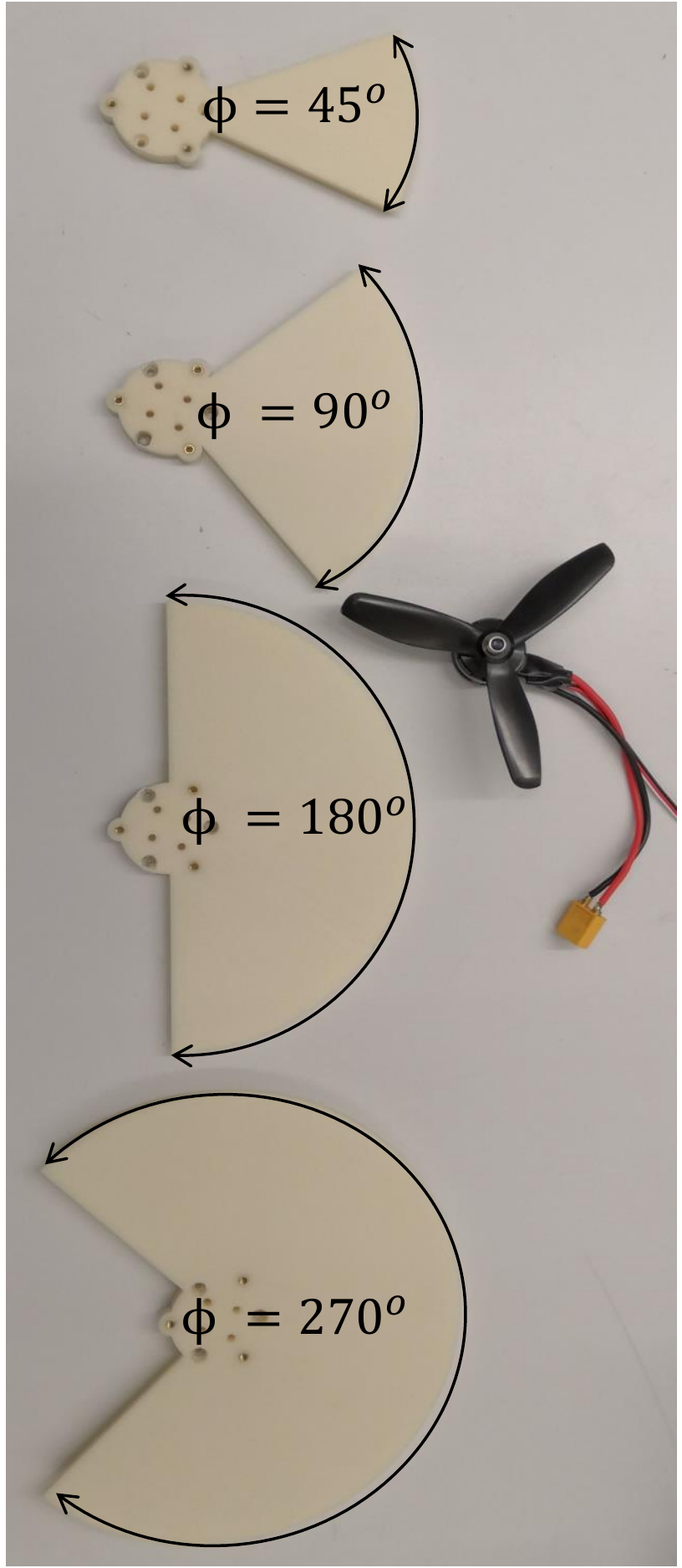}
    \caption{Partial\\ Occlusions}
    \label{fig: partial}
    \end{subfigure}%
  }\cr
  \noalign{\hspace{5mm}}
  \hbox{%
    \begin{subfigure}[b]{.23\textwidth}
    \centering
    \includegraphics[width=\textwidth]{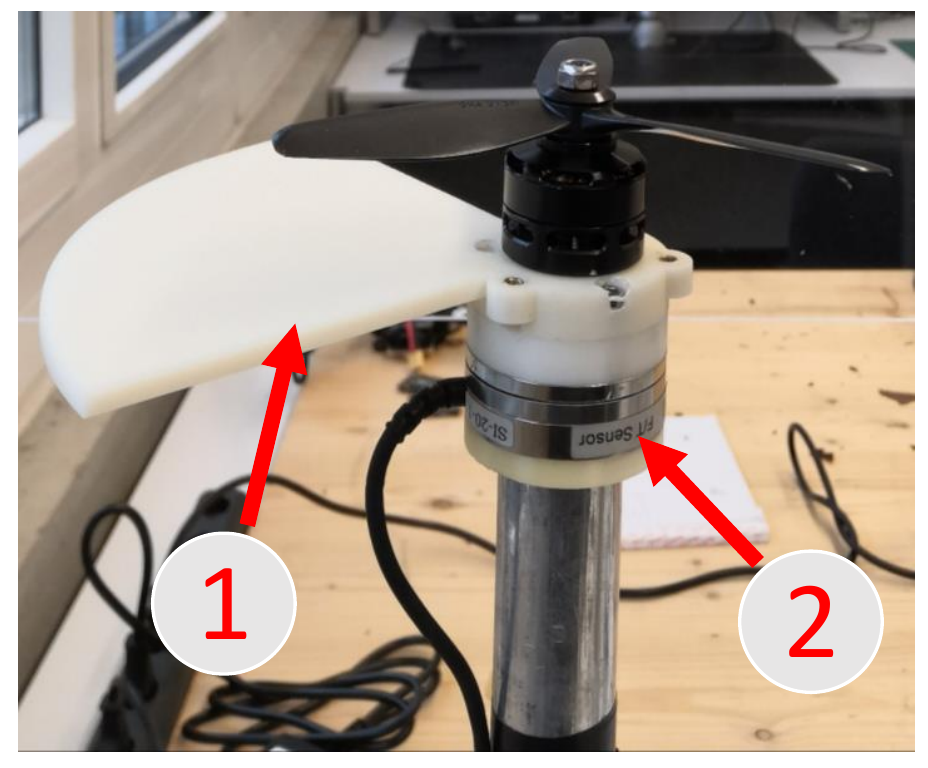}
    \caption{Load cell}
    \label{fig: load_cell}
    \end{subfigure}%
  }\vspace{4.8mm}
  \hbox{%
    \begin{subfigure}[b]{.23\textwidth}
    \centering
    \includegraphics[width=\textwidth]{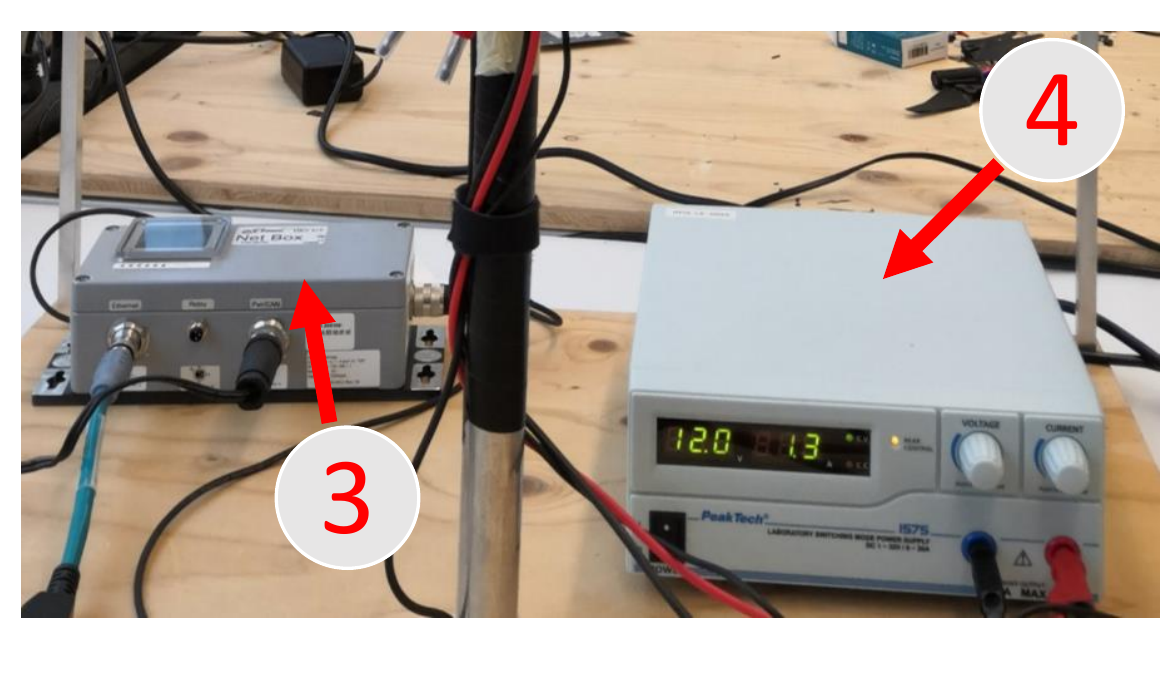}
    \caption{Netbox and PSU}
    \end{subfigure}%
  }\cr
}

\caption{Experimental setup. (a): 3D printed partial occlusions with 10cm radius replicating the Foldable Drone scenario; (b): $90\degree$ partial occlusion (1) fixed beneath the IQ motor on which is mounted the same propeller of the Foldable Drone and ATI Mini40 load cell (2); (c): (3) ATI Industrial Automation Net Box and (4) Power Supply Unit (PSU).}
\label{fig: exp_Setup}
\vspace{-2mm}
\end{figure}

In this paper, we propose and validate a simple framework to model the aerodynamic effects due to the \textit{partial overlap} between the propellers and the central body of a morphing quadrotor.
To this end, we model such interaction through experiments where the airflow of the propeller is hindered by different 3D-printed occlusions (cf. Fig.~\ref{fig: exp_Setup}). % resembling the scenario of the Foldable Drone when more compact configurations are assumed. 
This characterization is executed offline using a load cell to determine the parameters that link the overlap between the propeller disk and the central body to the loss of thrust with respect to the nominal case with no overlap.

As a case study, we consider the foldable drone proposed in~\cite{falanga2018foldable} in its most compact shape, the ``O" configuration (Fig. \ref{fig: O_config}), where all the arms are folded around the central body of the drone.
In this scenario, the drone experiences a loss of thrust, which makes it deviate from the reference position. 
%Nevertheless, the drone does not fall on the ground thanks to the low level controller of~\cite{falanga2018foldable} which, unaware of these aerodynamic effects, can only keep the position errors constant along all directions.
To counteract this issue, we quantify the impact of the overlap by comparing the results obtained with overlap to the nominal case with no occlusion underneath the propeller (e.g. the ``X" configuration Fig. \ref{fig: X_config}).
Additionally, we derive a real-time, geometry-aware formulation to compensate for these effects which does not rely on state information as in~\cite{zhao2017whole, zhao2018design}, and allows fast changes in configuration, more robust and stable control, and safer task completion. 
%This formulation makes the vehicle aware of its geometry and able to estimate the occlusions below the propellers online.
Finally, we deploy our geometry-aware compensation scheme on a real, fully autonomous morphing drone
%based the results of our performance analysis and characterization of our morphing drone, 
and show a significant increase in its tracking performance both in hovering and while tracking a reference trajectory.

\subsection{Related Work}
Most of the current morphing aerial platforms~\cite{riviere2018agile, bai2019evaluation, zhao2017whole, zhao2018design, bucki2019design} are either designed to perform single tasks,\iffalse completely relying on the effort of the low-level controller,\fi or based on passive morphing during flight.
The works in~\cite{riviere2018agile, bucki2019design, bai2019evaluation} present morphing quadrotors with different designs, but with the same limitation: they can only traverse vertical gaps. 
Both the systems of Riviere \etal and Bucki \etal traverse the gaps passively employing pre-computed trajectories. 
The former faces the gap at full speed while losing the controllability on the roll axis and leveraging accurate trajectory tracking thanks to an external motion capture system, whereas the latter relies on a ballistic motion where at its apex the gap is traversed. 
The last work guarantees continuous stable flight when its morphology is changed and the gap is being traversed, but at the cost of reduced dynamic stability in roll and pitch dynamics.
The works in~\cite{zhao2017whole, zhao2018design} propose the ``transformable multirotor" and the ``aerial robot dragon", respectively.
% \begin{figure}[t]
%     \centering
%     \includegraphics[width=0.5\textwidth]{img/front_page_image.pdf}
%     \caption{The Foldable Drone of~\cite{falanga2018foldable} flying in a circle trajectory at constant height and changing shape from X configuration to O configuration. Red trajectory: before the geometry-aware compensation scheme; Green trajectory: after deploying the geometry-aware compensation scheme.}
%     \label{fig: fig1}
% \end{figure}
The first design was mainly conceived to grasp objects, while the second one requires a large space before and after the traversal of all types of gaps. 
Moreover, both aerial vehicles are cumbersome and their complex maneuvers are time-consuming. 
On the other hand, the Foldable Drone proposed in~\cite{falanga2018foldable} provides a more versatile and flexible frame that can cross both vertical and horizontal gaps while being compact, and in stable flight in any configuration, even in asymmetric ones.\footnote{\url{https://youtu.be/jmKXCdEbF_E}} % due to their PID-based control algorithms which lowers their possibilities in a real life application. 
% Its design consists of four independently rotating arms that fold around the main frame, and a control scheme that guarantees stable flight also in asymmetric configurations.
%We therefore choose to characterize the Foldable Drone, and analyze its performance.

All aforementioned works ignore most of the thrust losses generated by the interaction between the propellers and the rest of the vehicle's body. For instance, aerodynamic effects have been analyzed in literature only for fixed-shape quadrotors.
In~\cite{nandakumar2018theoretical} the authors presented theoretical and experimental investigations of a Vertically Offset Overlapped Propulsion System (VOOPS) where only the overlap between two adjacent propellers was considered. The most beneficial amount of overlap between propellers was determined offline, and the partial overlap between a propeller and the central body of the drone was not analyzed.
The work in~\cite{yoon2017computational} analyzed in simulation the aerodynamic interactions in fixed shape quadrotors showing the ubiquitous loss of thrust due to the overlap between the propellers and the arms of a quadcopter.
These results agree with~\cite{theys2016influence} and~\cite{fernandes2011design}, where the efficiency of the system with different arms mounted beneath different propellers was analyzed.
However, this phenomenon may be accentuated in a morphing frame according to the dimensions of the regions of high pressure.
Indeed, these issues not only depend on the regions of high pressure that are generated whenever the blades of the propeller overlap with the arms of the drone, but also on the different types of occlusions that hinder the wake of the propeller. The consequent loss of thrust is always present in quadrotors with upward mounted propellers, whereas, as shown in~\cite{pounds2004towards, theys2016influence}, that is not the case for quadcopters with downward mounted rotors.% which are claimed to have better performance.
% Finally, in~\cite{hooi2015flow,sanchez2017characterization,bernard2017dynamic} the ground effect in multirotors was analyzed.
% This aerodynamic effect is present in all rotor-based aerial vehicles and consists of the increment of thrust generated by the rotors when flying close to the ground due to the interaction of the rotor airflow with the ground surface. 
% Sanchez \etal~\cite{sanchez2017characterization} started from an analytical model of the ground effect in helicopters and analyzed what they called "the partial ground effect", a situation in which only one or some of the rotors of the multirotor (but not all) are under the ground effect. Their analysis did not consider the propeller-body overlap which may happen in case of high proximity between the rotors and the robotic arm beneath them.
% The work in~\cite{hooi2015flow} provided a nonlinear dynamic model of the rotorcraft in ground effect using multiple ring sources. 
% % Their work was an extension of the classic ground effect based on the method of images.
% Nevertheless, this model seemed to provide a more accurate characterization of the ground effect only for very large propellers.

% \begin{figure}[!t]
% \captionsetup[subfigure]{justification=centering}
%      \centering
%      \begin{subfigure}[t]{0.23\textwidth}
%          \centering
%          \includegraphics[width=\textwidth]{img/loadcell_occl_cropped.pdf}
%          \caption{Load cell}
%          \label{fig: load_cell}
%      \end{subfigure}
%      \begin{subfigure}[t]{0.23\textwidth}
%          \centering
%          \includegraphics[width=\textwidth]{img/netbox_psu_cropped.pdf}
%          \caption{Netbox and PSU}
%          %\label{fig: load_cell}
%      \end{subfigure}
%      \hspace{5mm}
%      \begin{subfigure}[t]{0.1365\textwidth}
%          \centering
%          \includegraphics[width=\textwidth]{img/partial.pdf}
%          \caption{Partial\\ Occlusions}
%          \label{fig: partial}
%      \end{subfigure}
%     \caption{Pictures of our experimental setup: (a) ATI Mini40 load cell and Power Supply Unit (PSU) that we used for all our measurements; (b) the 3D printed partial occlusions having radius of 10cm which replicate the Foldable Drone scenario.}
%     \label{fig: exp_Setup}
% \end{figure}
\vspace{-1mm}
\subsection{Structure of the Paper}
In the following, we give an overview of the organization of the paper.
In Sec. \ref{sec:exp}, we present the experiments regarding the partial overlap. In Sec. \ref{sec:comp}, we reveal the formulation of the geometry-aware compensation scheme. In Sec. \ref{sec:dep}, we demonstrate the effectiveness of our formulation against~\cite{falanga2018foldable}. Finally, in Sec. \ref{sec:conc} and \ref{disc}, we illustrate the conclusions.
\section{Methodology}\label{sec:exp}
In this section, we motivate our method, we illustrate the adopted experimental setup for the identification of our model, we provide a geometric description of the tested occlusions and we discuss the experimental results along with their physical implications.
\subsection{Motivation}
We propose a simple, yet effective, identification for our model which leverages an experimental approach. However, a more theoretical approach could be followed by, for example, analyzing the nonlinear aerodynamic effects generated by the interaction between the airflow of the propellers and the presence of different partial occlusions. Although the second method might lead to a more precise solution, its derivation would be unfeasible for a real-time system, and it would also require an experimental parameter identification. The simplicity of our method allows for a more general formulation of the problem while achieving precise results and allowing real-time computation of the necessary parameters.
\subsection{Model Formulation}
The thrust $T$ produced by a single propeller is given by~\cite{leishman} as:
\vspace{-2mm}
\begin{align}\label{eq: eq1}
T= \rho A C_t(R\omega)^2,
\end{align}
where $\rho$ is the density of air, $A$ is the area of the propeller disk, $C_t$ is the coefficient of thrust, $R$ is the radius of the propeller and $\omega$ is the angular velocity of the propeller.
In this paper we will simplify this notation as follows:
\vspace{-2mm}
\begin{align}\label{eq: T_form}
T= K_T\omega^2
\end{align}
where $K_T$ is what we call the thrust coefficient, which encloses all the constants as explicitly expressed in Eq. \ref{eq: eq1}, and $\omega$ is the angular velocity of the propeller. This notation corresponds to what we define as the nominal case i.e. the case of no occlusions beneath the propeller.
To take into account the presence of a partial occlusion beneath the rotor we reformulate Eq. \ref{eq: T_form} as:
\vspace{-2mm}
\begin{align}\label{deg_dep_coeff}
T= k(\phi)\omega^2,
\end{align}
where $\phi$ is the angle that characterize the different partial occlusion and $k(\phi)$ is the related coefficient. 
We will refer to this coefficient as the \textit{angle-dependent coefficient}. It is important to note that, in our notation, $k(\phi=0\degree)$ corresponds to $K_T$ which is measured with the same experimental setup and no occlusion beneath the propeller.\\
\vspace{-3mm}
\subsection{Experimental Setup for Parameters Identification}
Our experimental setup consists of an ATI Mini40 load cell (cf. Fig. \ref{fig: load_cell}) which allows us to measure the thrust produced by the propeller with and without the presence of an occlusion beneath it. 
On top of the sensor, we mounted the occlusions depicted in Fig. \ref{fig: partial} and the IQ2306 Speed Module 2200KV, an integrated motor and controller which provides feedback on the velocity of the rotor.
Thanks to the feedback provided by the motor and the thrust measurements we can precisely determine the thrust coefficient of the propeller with different geometric occlusions and at different speeds. On top of the motor, we singularly tested $5$ inches propellers ($2$ and $3$ blades) and $6$ inches propellers ($2$ and $3$ blades). 

In all the experiments, we provided a voltage of 12V to the motor through the PSU.
To explore widely the trend of the thrust with different occlusions, we gave as input to the motor an angular velocity consisting of $40$ steps of $2$ seconds each from a minimum of $100 \hspace{1mm} RPM$ to a maximum of $19100 \hspace{1mm} RPM$.
Due to the physical limitations of the motor, an exception was made for the $3$ blades, $6$ inches propeller, in which case a maximum motor speed of $17188 \hspace{1mm} RPM$ was provided. It is important to note that, the thrust loss with different partial occlusions and low motor speed is still present. \iffalse However, in the following, we only present the results obtained with the aforementioned maximum motor speeds as they replicate the airflow conditions of a normal flight.\fi

In Fig. \ref{fig: partial}, we show one of the novelties of this work, that is, the tested occlusions that partially hinder the airflow of the propeller to replicate the Foldable Drone scenario. 
They consist of different 3D printed sectors of a fixed size disk characterized by different angles $\phi \in \{45 \degree,90\degree,180\degree,270\degree \}$ which are designed to be mounted on top of the load cell and below the motor. 
In this paper, we will refer to these occlusions as \textit{partial occlusions}.

\begin{figure}[!t]
     \centering
     \includegraphics[width=0.35\textwidth]{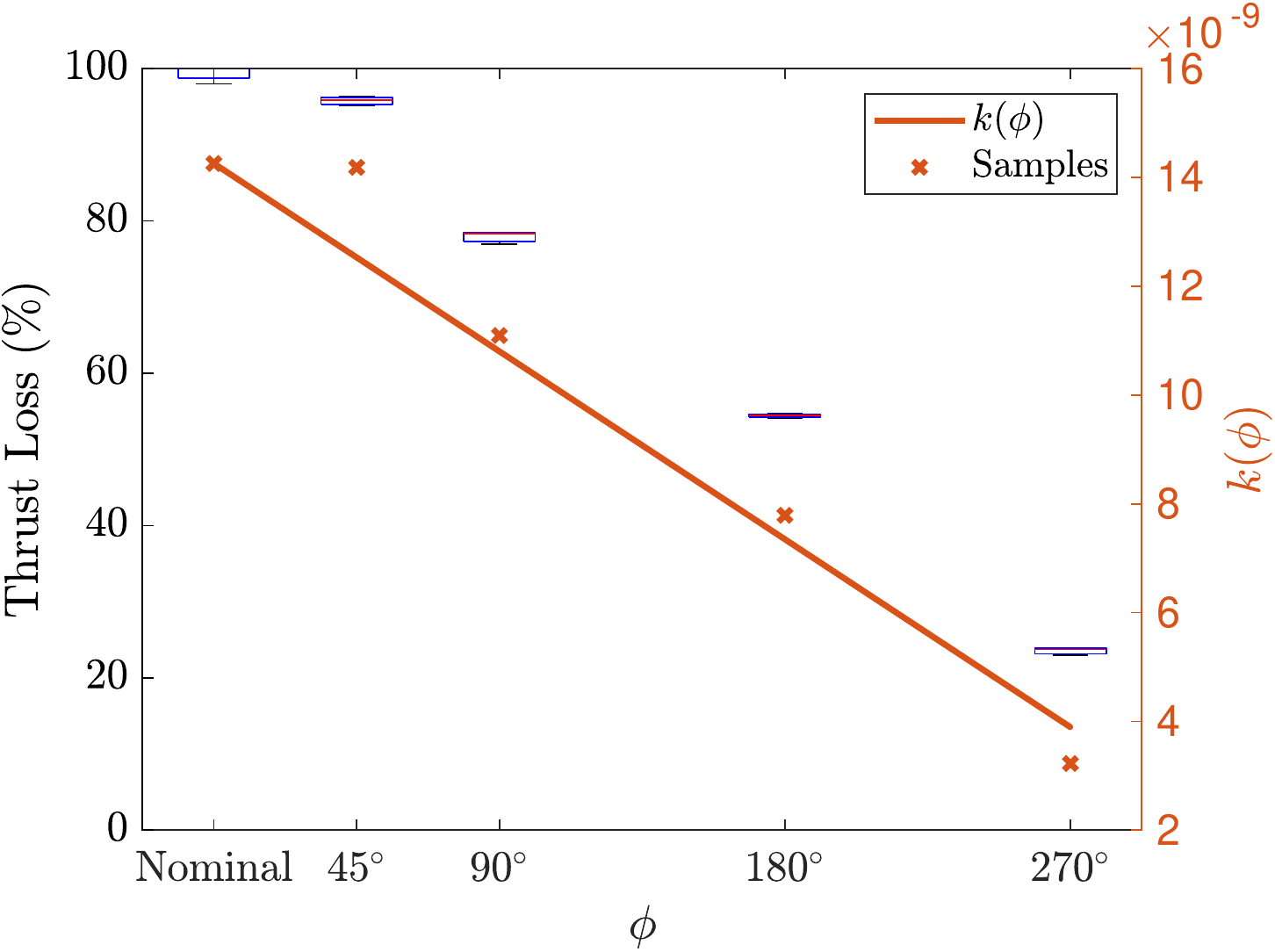}
    \caption{Boxplot of $3$ blades $5$ inches propeller showing the loss of thrust with respect to the nominal case due to the partial occlusions (x-axis: $\phi$). In solid orange, the fitting of the angle-dependent coefficient on our experimental data. Both plots are obtained with the same propeller of the Foldable Drone spinned at $19100\hspace{1mm}RPM$.}
     \label{fig: T_part}
    \vspace{-2mm}
\end{figure}
\subsection{Results of the Identification}
\begin{figure}[!t]
     \centering
     \begin{subfigure}[t]{0.23\textwidth}
         \centering
         \includegraphics[width=\textwidth]{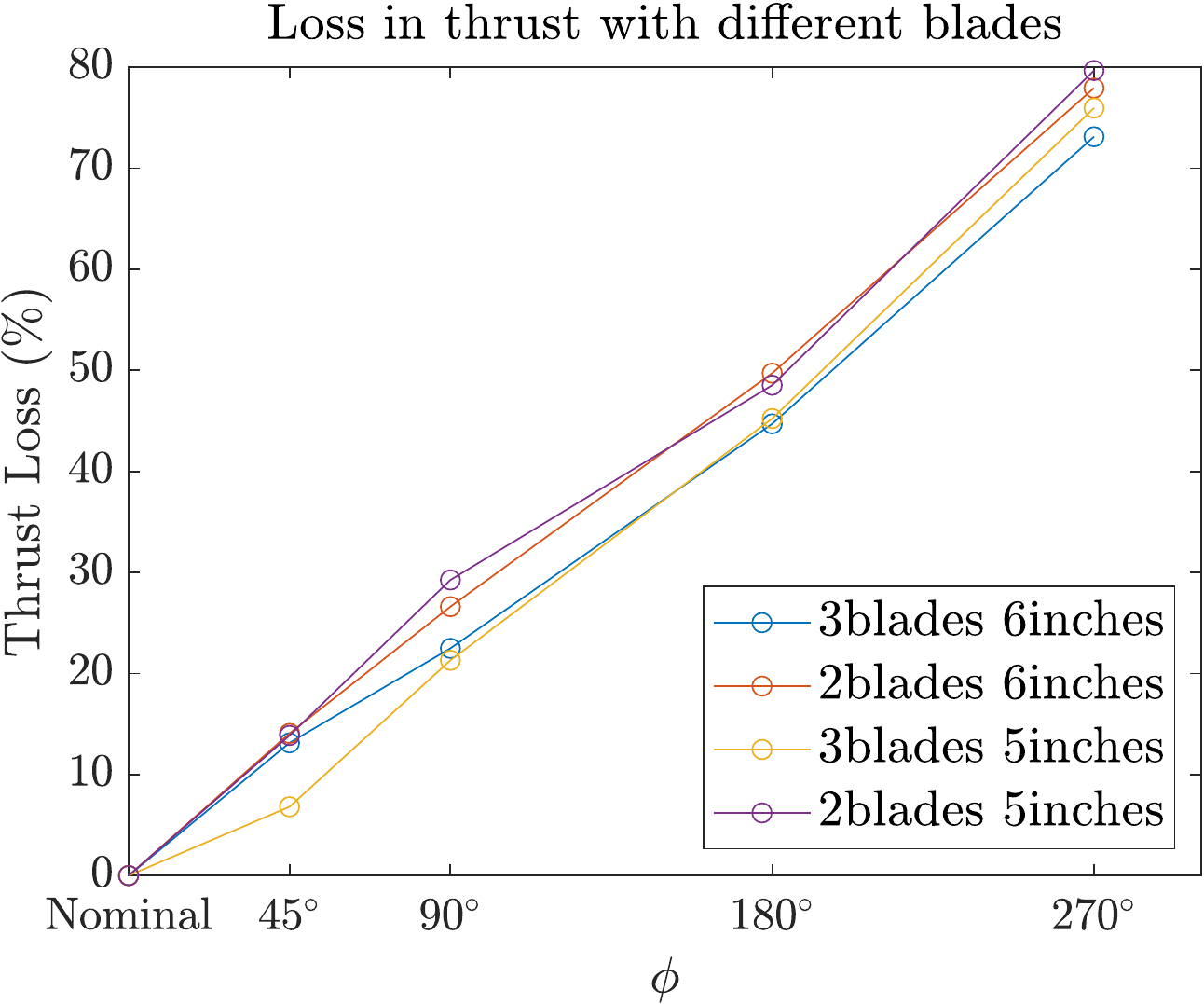}
         \caption{Thrust loss percentage for different propellers.}
         \label{fig: all_blades_T}
         \vspace{2mm}
     \end{subfigure}
     \hfill
     \begin{subfigure}[t]{0.23\textwidth}
         \centering
         \includegraphics[width=\textwidth]{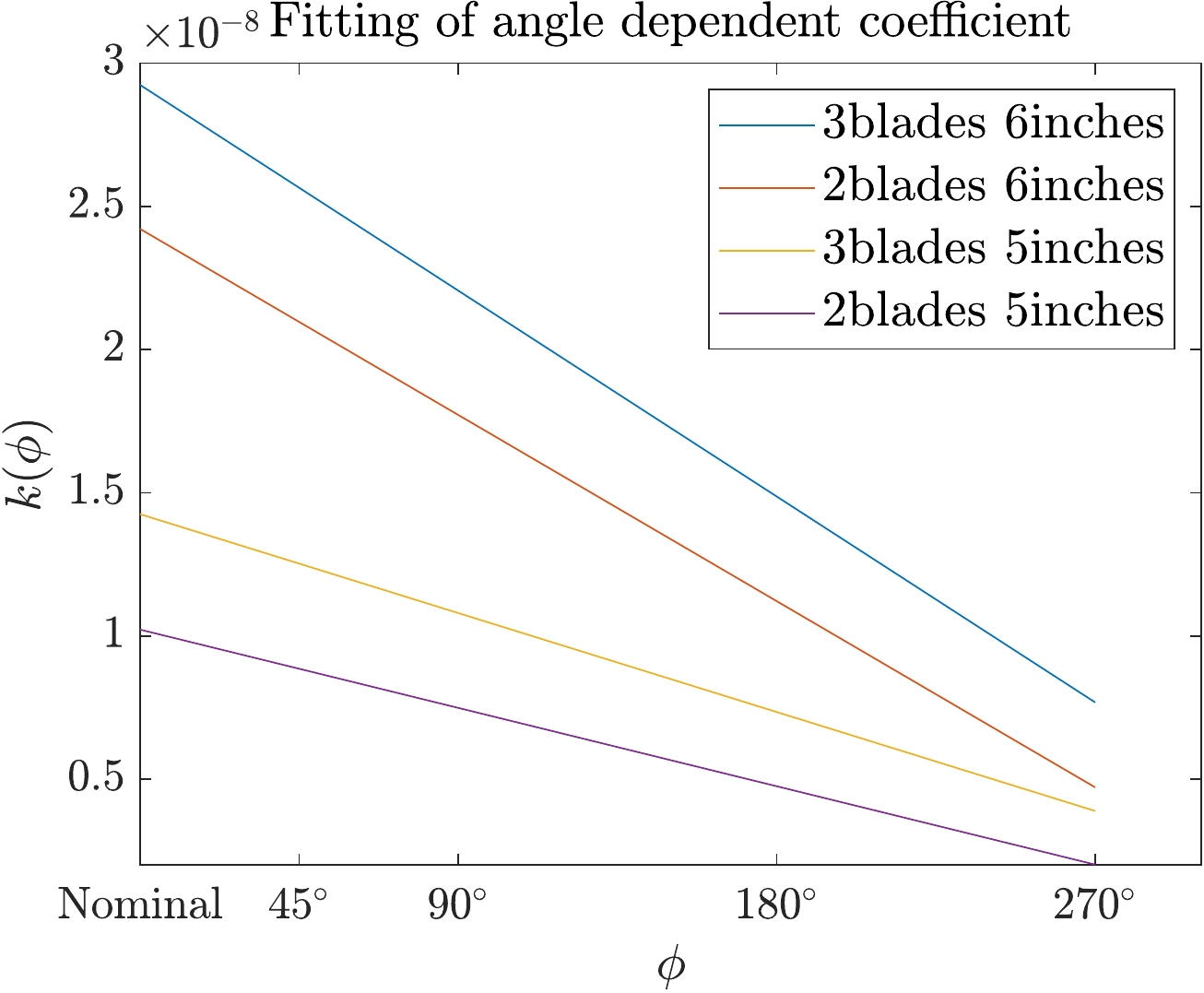}
         \caption{The angle-dependent thrust coefficient for different propellers.}
         \label{fig: all_blades_k}
     \end{subfigure}
    \caption{Plot highlighting the comparison between different propellers: (a) the loss of thrust with respect to the nominal case due to the partial occlusion (all tested propellers); (b) the angle-dependent coefficients of all the tested propellers. Both plots refer to the input $17188\hspace{1mm}RPM$.}
    \label{fig: all_blades}
    \vspace{-2mm}
\end{figure}

For each partial occlusion, we observe how the thrust and the related thrust coefficient, at different motor speeds, change as a function of the angle of occlusion $\phi$. 
We consider the thrust loss at the nominal case, i.e. no occlusion, as a function of the angle of the partial occlusions $\phi$.
As expected, the loss of thrust is larger with a growing angle, that is, with bigger occlusions. Our experiments show that the thrust produced goes down linearly (Fig. \ref{fig: T_part}). For the case of the $3$ blades $5$ inches propeller, that is the propeller that is mounted on the Foldable Drone, the produced thrust ranges from almost $90\%$ to $25\%$ of the nominal thrust with partial occlusions of $\phi=45\degree$ and $\phi=270\degree$ respectively. This is because of the corresponding angle-dependent coefficient $k(\phi)$ of Eq. \ref{deg_dep_coeff} follows the same trend (Fig. \ref{fig: T_part}).
In Fig. \ref{fig: all_blades} we can see the comparison between different propellers on the same occlusions. The general trend is repeated uniformly with the opportune scale of the thrust due to the different dimension of the propeller and its number of blades.

Efficiency being the ratio of input power to obtained lift, 2 blades propellers are more efficient than 3 blades propellers. Nevertheless, the last two produce more thrust than the first two if the partial occlusion is big enough ($\phi>45\degree$, cf. Fig. \ref{fig: all_blades_T}). 
Indeed, the more thrust a rotor produces, the bigger the occlusion needs to be to hinder the propeller in analysis. 
In Fig. \ref{fig: all_blades_k}, a correlation between the angle-dependent coefficients of different propellers can be deducted for big partial occlusions. For the case of propellers with the same diameter, but different number of blades, the slopes of the angle-dependent coefficients can be considered to be the same, whereas the vertical offset between the two lines increases with the dimension of the propeller, $0.5\times10^{-8}$ between the $6$ inches propellers and $0.4\times10^{-8}$ between the $5$ inches propellers. Moreover, for the case of propellers with different dimensions, but same number of blades, the value of the angle-dependent coefficient doubles if the difference between their diameters is $1$ inch. These experiments allow us to generalize the estimation of the angle-dependent coefficient to a broad variety of morphing and non-morphing aerial vehicles with different propellers.
It is important to note that, in the Foldable Drone scenario, a bigger occlusion corresponds to a larger overlap between the propellers and the central body of the drone. Moreover, for the case of our morphing quadrotor, partial occlusions with $\phi > 65\degree$ are not present due to the mechanical design of our platform.

Since all the measurements are assumed to be affected by white noise, we counteract this issue with averaging. Thus, we repeated each measurement three times and took the average of these three repetitions. The consequences of this operation are a more reliable approximation of the measurements with high variance and a more reliable compensation scheme.

\section{Geometry-Aware Compensation Scheme}\label{sec:comp}
In this section, we provide the assumptions and the model approximations upon which our compensation scheme is based on along with its formulation.

\subsection{Assumptions}
\begin{figure}[!t]
     \centering
     \begin{subfigure}[t]{0.2\textwidth}
         \centering
         \includegraphics[width=\textwidth]{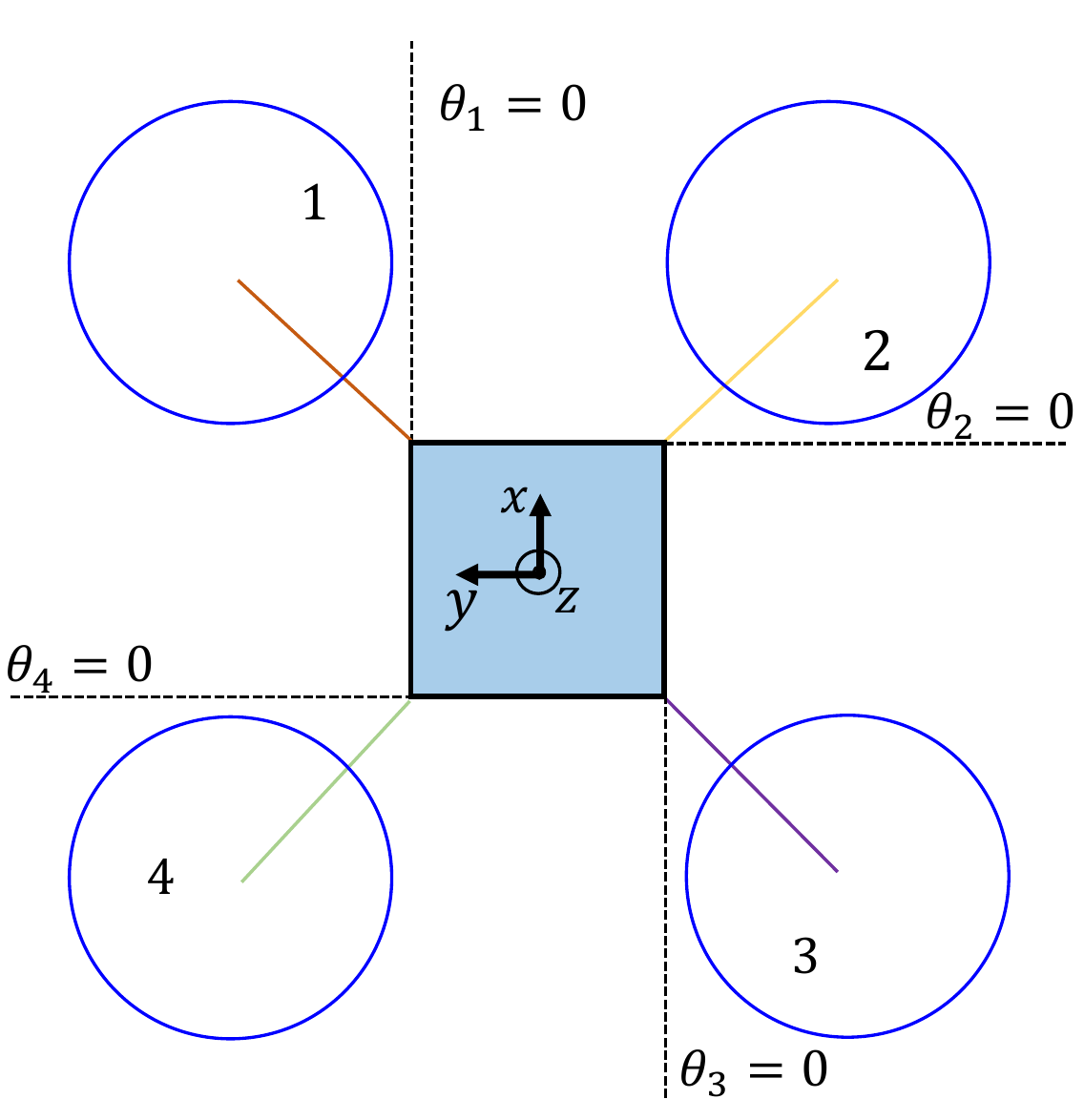}
         \caption{X configuration: no overlap.}
         \label{fig: X_config}
     \end{subfigure}
     \hspace{5mm}
     \begin{subfigure}[t]{0.2\textwidth}
         \centering
         \includegraphics[width=\textwidth]{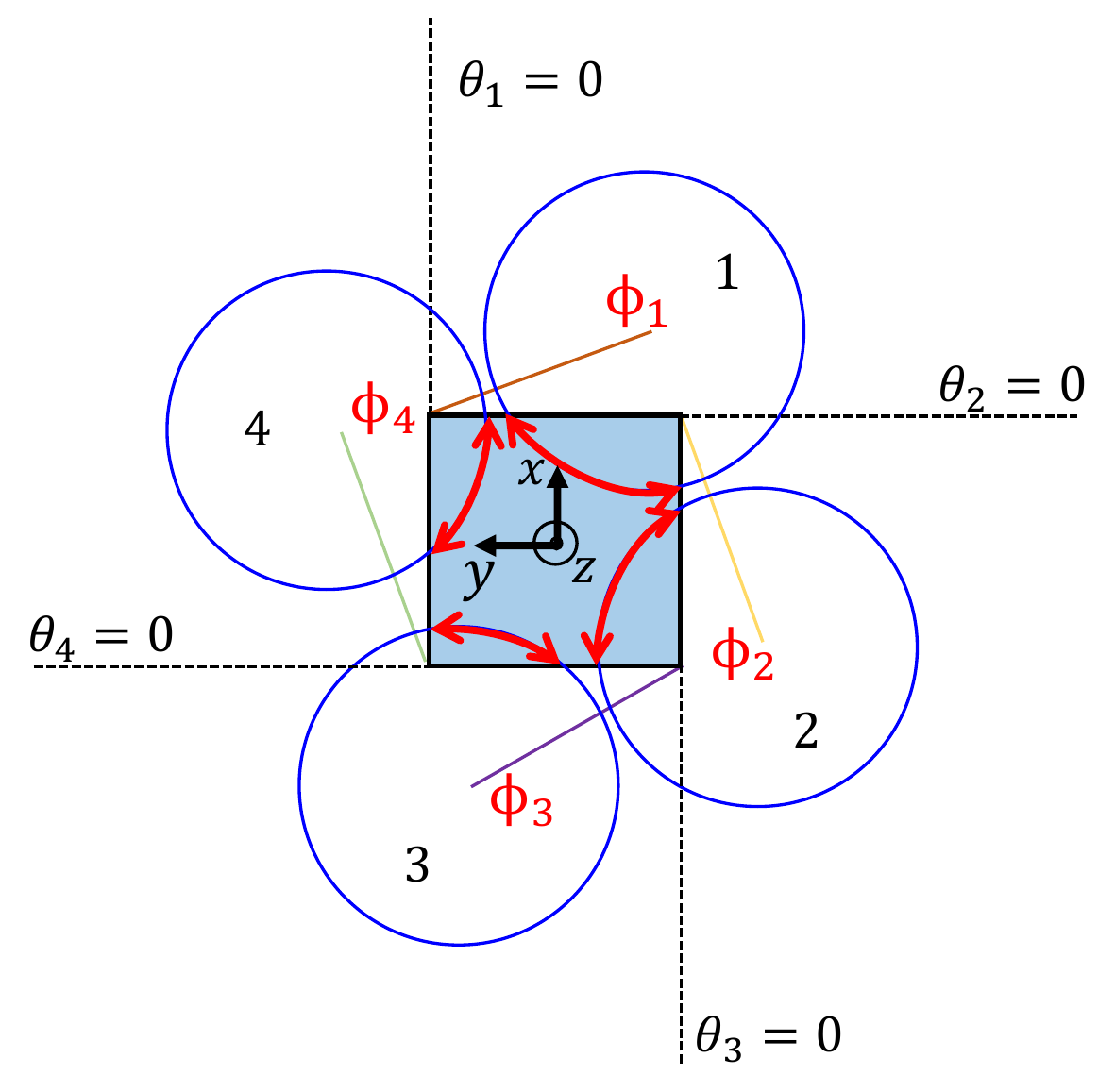}
         \caption{O configuration: overlap.}
         \label{fig: O_config}
         \vspace{2mm}
     \end{subfigure}
    %  \hfill
    %  \begin{subfigure}[t]{0.3\textwidth}
    %      \centering
    %      \includegraphics[width=\textwidth]{img/3b_5inches_FittingCoefficientDegreeDependentPartial.pdf}
    %      \caption{Fitting of the angle-dependent coefficient.}
    %      \label{fig: T_part}
    %  \end{subfigure}
    \caption{Schematics showing the default configuration of the Foldable Drone (a) and its most compact configuration (b) highlighting the angle of overlap between the propellers and the central body of the drone. \iffalse On the bottom the fitting of the angle-dependent coefficient on our experimental data obtained with the same propeller of the Foldable Drone spinned at $19100\hspace{1mm}RPM$.\fi}
    \label{fig:configs}
    \vspace{-4mm}
\end{figure}

Because, for the case of the Foldable Drone, the overlap between propellers is less than $25\%$ (for example in ``H" and ``T" configurations), as suggested in~\cite{nandakumar2018theoretical}, the consequent loss of thrust is less than $5\%$, and hence negligible. In~\cite{falanga2018foldable} the loss of thrust starts to be relevant whenever an overlap between the propellers and the central body of the vehicle occurs, due to the formation of a region of high pressure. Furthermore, the geometry-aware compensation scheme assumes that the main contribution of the thrust is near the tips of the rotor due to the high-velocity airflow~\cite{yoon2017computational}. Finally, small distances of a maximum of 2 cm plus the height of the motor from the partial occlusions are ignored as they lead to negligible or absent gain of thrust.\\
To estimate geometrically the partial occlusions, we simplified the quadrotor as follows (Fig. \ref{fig: X_config}): the central body of the drone is considered to be a square, and the arms to be 4 dimension-less lines. We denote the angles of the arms rotated with respect to the central body as $\theta_i$ while we denote the angles that characterize the partial occlusions as $\phi_i$ where $i \in \{1,2,3,4\}$. Each arm can rotate in the following range $\theta_i \in [-\frac{\pi}{2};\pi]$. In this work, we will refer to $\phi_i$ as the angle of occlusion. It is important to note that the angle of occlusion is computed at each time the drone decides to change its shape while flying by only knowing $\theta_i$, and the geometry of the vehicle. Moreover, if there is no overlap between the propellers and the central body of the drone, the thrust mapping remains unchanged.
% \begin{figure*}[!t]
% \begin{center}
%     \includegraphics[width=0.32\textwidth]{img/XO_plot_bound_no_comp.pdf} % first figure itself
%     \includegraphics[width=0.3\textwidth]{img/XO_plot_bound_comp.pdf} % first figure itself
%     \includegraphics[width=0.32\textwidth]{img/O_plot_bound_comp_on_off.pdf}
% \end{center}
%     \caption{Plots of the position error for the experimental validation of our geometry-aware compensation scheme. Left: transition in hover from X to O configuration without compensation. Center: transition in hover from X to O configuration with compensation. Right: hovering in O configuration switching on and off the compensation. Each row corresponds to a different axis of the world frame.}
%     \label{img:results_hover}
%     \vspace{-4mm}
% \end{figure*}

\subsection{Formulation of the Compensation Scheme}
The geometry-aware compensation scheme aims to modify the motor speeds of the Foldable Drone to overcome the effect of overlap while maintaining the same desired thrust of the previous configuration. To avoid the loss of thrust due to the overlap between the propellers and the central body of the drone, the angular velocities of the rotors must increase. To this end, we propose the following formulation for each propeller $i=\{1,2,3,4\}$:
\begin{align}
    \label{eq:4}
    \hat{\omega}_i = \sqrt{\frac{T_i}{K_Tk_i}}, \hspace{5mm}
    k_i = 1 - \frac{|K_T - k(\phi_i)|}{K_T}
\end{align}
where $\hat{\omega}_i\in \mathbb{R}$ is the modified desired motor speed of the i$^{th}$ propeller which counteracts the loss of thrust, $T_i\in \mathbb{R}$ is the thrust intensity of the i$^{th}$ propeller that we want to produce, and $k_i\in \mathbb{R}$ is a scaling factor of the i$^{th}$ propeller which depends on the angle-dependent coefficient of the i$^{th}$ propeller $k(\phi_i)\in \mathbb{R}$, and on what we called the thrust coefficient $K_T\in \mathbb{R}$. From our experimental data in Sec. \ref{sec:exp}, knowing that the trend of the angle-dependent coefficient as a function of different partial occlusions is linear with respect to their angles, we can fit a line to our data with the least-squares method (cf. Fig. \ref{fig: T_part}). Once we geometrically derive $\pmb{\phi}$, that is, the angle of occlusion of all the overlapped propellers, we can determine their corresponding angle-dependent coefficients. The subtraction in Eq. \ref{eq:4} is related to the fact that the thrust decreases as the occlusions increase.
Hence, the $i^{th}$ scaling factor $k_i \in [0,1]$ needs to be smaller than one to increase the angular velocity of the $i^{th}$  propeller $\omega_i$.
From the formulation in Eq. \ref{eq:4}, it derives that the amount of angular velocity to overcome the effect of the occlusion is $100\sqrt{\frac{1}{k_i}}$. Indeed, if $k_i(\phi) = 0$ then $k_i=0$, the occlusion is infinitely big and the rotors must spin infinitely fast to overcome the effect generated by such occlusion.
On the other hand, if $k_i(\phi) = K_T$ then $k_i=1$, there is no occlusion (e.g. Fig. \ref{fig: X_config}) and consequently the rotors can spin as fast as before.
Nevertheless, the motor might saturate in trying to overcome the loss of thrust introduced by the presence of a very big occlusion. In~\cite{falanga2018foldable}, this problem is already taken into account. We apply the saturation scheme for the rotor thrusts proposed in~\cite{faessler2016thrust}, which prioritizes between the desired collective thrust and body torques according to their relevance for stabilizing the quadrotor and following a trajectory. Moreover, the motors of the Foldable Drone are controlled by a Qualcomm Snapdragon Flight closed-loop ESC. Therefore, the correct thrust mapping during the discharge of the battery is provided.
Finally, our geometric-aware compensation scheme can handle also nonsymmetric configurations of the Foldable Drone since it computes the angle-dependent coefficient for each propeller independently. The computation of the angles of occlusion, and their corresponding angle-dependent coefficients, is performed online, but only when a specific geometric condition, that checks the overlap between a propeller and the central body of the quadcopter, is met.

\section{Deployment and Results}\label{sec:dep}

In this section, we present and discuss the performance of our proposed compensation scheme.
We deployed our scheme on the Foldable Drone and analyzed the performance of the quadrotor in nine experiments during both hovering and forward flight with vision-based navigation.
All the experiments are made both with compensation (c.) and no compensation (n.c.).
\\
\textbf{Hovering} - The drone hovers in X configuration. The robot changes its shape during hovering from X to O configuration which is the most compact one and the most affected by the loss of thrust due to overlap between the propellers and the central body of the drone.
When the drone ends the transition and reaches stable hovering it goes back to the X configuration.
We performed the aforementioned experiments with and without compensation scheme.
Finally, the morphing quadrotor always hovers in O configuration while switching on and off the compensation scheme.\\
\textbf{Forward Flight} - The drone follows a circular trajectory at a speed of 0.6 m/s on a circle of radius 1.5 m, at a constant height of 1.5 m. The robot changes its shape during flight switching between X and O configurations with and without the activation of the compensation scheme. Finally, the Foldable Drone flies always in O configuration while switching on and off the compensation scheme.
The same aforementioned experiments are then performed with the drone following a circular trajectory at a speed of 0.6 m/s on a  circle of radius 1.5 m, but this time with a varying height that ranges from a minimum of 1.25 m to a maximum of 1.75 m.
Tab. \ref{ERR_pose_h} and  \ref{ERR_pose_ff} show some quantitative results that prove the effectiveness of our proposed compensation.

\begin{table}[t!]
\begin{center}
\begin{tabular}{|c|c|c|c|c|}
  \hline
  \multicolumn{5}{|c|}{\tabincell{c}{Experiments: Transitions in Hovering}}\\
  \hline
 \tabincell{c}{Start \\ config} & \tabincell{c}{Final \\ config}
  & \multicolumn{3}{c|}{Position errors in final config [m]}  \\
  \cline{3-5}
  & &  \tabincell{c}{$|x_{r}-x_{e}|$}  & \tabincell{c}{$|y_{r}-y_{e}|$} &  \tabincell{c}{$|z_{r}-z_{e}|$} \\
  \hline
  X & X & 0.0336 & 0.0185 & 0.0142\\
  \hline
  \hline
  X & O$_{n.c.}$ & 0.22 & 0.11 & 0.17\\
  \hline
  X & \textbf{O}$\mathbf{_{c.}}$ & \textbf{0.030} & \textbf{0.044} & \textbf{0.043}\\
  \hline
  \hline
  O$_{n.c.}$ & X & \textbf{0.040}
& 0.049 & 0.031\\
  \hline
  \textbf{O}$\mathbf{_{c.}}$ & X & 0.047 & \textbf{0.042} & \textbf{0.00855}\\
  \hline
  \hline
   O$_{c.}$ & O$_{n.c.}$ &0.0748 & 0.0222 & 0.1594 \\
  \hline
  O$_{n.c.}$ & \textbf{O}$\mathbf{_{c.}}$ & \textbf{0.0101} & \textbf{0.0003} & \textbf{0.0106}\\ 
  \hline
\end{tabular}
\caption{Position errors of the Foldable Drone computed as the absolute difference between the reference state (desired state) and the state estimate (approximation of the actual state) of the vehicle. The errors refer to the final configuration of the transition in hovering. In bold the best results.\iffalse ``c." stands for compensation and, ``n.c." stands for no compensation.\fi}
\label{ERR_pose_h}% SOTA with Test
\end{center}
\vspace{-4mm}
\end{table}
\subsection{Results}
The compensation scheme improves the position tracking both in hovering and forward flight when the most compact configuration (i.e., O) is assumed.
If the drone always hovers in X configuration the position errors are below 3.5 cm.
When the drone switches to the O configuration and no compensation is provided, the quadrotor drifts more than 10 cm away from the desired position along all directions.
For instance, the drone drops down approximately 17 cm below the reference state and shifts 21 cm and 11 cm away from it along the x-axis and y-axis respectively (Tab. \ref{ERR_pose_h}). 
This drop (Fig. \ref{fig: X_to_O_nocomp}) is due to the loss of thrust generated by the overlap between the propellers and the central body of the Foldable Drone.
The experiments highlight also a translation along the $x$ and $y$ directions.
% \begin{figure*}[t]
% \begin{center}
%     \includegraphics[width=0.32\textwidth]{img/X_to_O_no_comp_traj_circle_h_const.pdf}
%     \includegraphics[width=0.3\textwidth]{img/X_to_O_comp_traj_circle_h_const.pdf}
%     \includegraphics[width=0.32\textwidth]{img/O_on_off_traj_circle_h_const.pdf}
% \end{center}
%     \caption{3D Plots of the circle trajectories at constant height flown by the Foldable Drone. The blue circles represent the reference trajectories. Left: in red O configuration without compensation; in green X configuration. Center: in red O configuration with compensation; in green X configuration. Right: in red O configuration without compensation; in green O configuration with compensation.}
%     \label{img:results_ff}
% \end{figure*}
On one hand, this translation derives from the induced motion given by the rotation of arms while the drone drops and the turbulent flows generated by the overlaps.
On the other hand, it is due to the fact that all the arms are not rotated by the same angle (cf. Fig. \ref{fig: O_config}).
Indeed, propeller 3 overlaps less with the central body than the other propellers. This is due to a geometric constraint related to the location of the battery on the Foldable Drone.
This means that propellers 1, 2, and 4 experience a stronger loss of thrust than propeller 3. This difference of thrusts makes the drone pitch and translate along the $x$-axis (21 cm away from the reference state) and slightly roll and translate along the $y$-axis (11 cm away from the reference state). The same transition in hovering from X to O configuration is much more stable when the compensation scheme is activated.
Thanks to the integration of our geometry-aware compensation scheme, the position errors in the O configuration are really close to the ones in the X configuration (all below 4.5 cm).
Finally, when the drone switches back to the X configuration, the desired state is reached with almost the same errors as before.
This is the only case where the compensation scheme does not improve significantly the position tracking of the drone apart from the error along the $z$-axis.

The last experiment in hovering does not comprehend any changes in the shape of the drone during flight. The Foldable Drone always hovers in O configuration. \iffalse During flight, we switch on and off the compensation scheme to demonstrate how the aerodynamic effect of the overlap is ignored if this is deactivated.\fi
When the compensation is activated, the position errors are again drastically reduced, whereas, when the compensation scheme is switched off, the opposite occurs.
In this case, the biggest difference in position errors is along the $z$-axis since no change in configuration takes place and consequently, no induced motion is produced.

Our compensation scheme boosted the performance of the morphing frame also in forward flight almost halving the euclidean distance between the reference trajectory and the actual one (cf. Tab. \ref{ERR_pose_ff}). The activation of the compensation scheme allows the drone to immediately compensate for the aerodynamic effect regardless of what the previous configuration was, and without losing the track of the trajectory or deviate from it significantly.

\begin{table}[t!]
\begin{center}
\begin{tabular}{|c|c|c|c|c|}
  \hline
  \multicolumn{5}{|c|}{\tabincell{c}{Experiments: Transitions in Hovering}}\\
  \hline
  \tabincell{c}{Start \\config} & \tabincell{c}{Final \\config}
  & \multicolumn{3}{c|}{$\Ddot{e} = k_Pe + k_D\Dot{e}$}  \\
  \cline{3-5}
  & &  \tabincell{c}{$\Ddot{e} \hspace{1mm}[^m/_{s^2}]$}  & \tabincell{c}{$e\hspace{1mm} [m]$} &  \tabincell{c}{$\Dot{e}\hspace{1mm} [^m/_s]$} \\
  \hline
  X & O$_{n.c.}$ & -0.50052
 & 0.16859 & -0.50489 \\
  \hline
  X & \textbf{O}$\mathbf{_{c.}}$ & \textbf{-0.03259}& \textbf{-0.0434} & \textbf{0.10318}\\
  \hline
\end{tabular}
\caption{Errors contribution to the PD controller for gravity compensation. We define $e := z_{ref} - z_{est}$. \iffalse Note: ``c." stands for compensation and, ``n.c." stands for no compensation.\fi}
\label{ERR_acc}% SOTA with Test
\end{center}
\vspace{-3mm}
\end{table}

To further understand the impact of our proposed compensation scheme, we compute its contribution to the already present PD controller for the gravity compensation (Tab. \ref{ERR_acc}).
When the Foldable Drone is hovering, the total acceleration should be equal to gravity. If the compensation is not activated, gravity acceleration is reached with a strong contribution of the position error. However, when the compensation is used, the opposite occurs. This thrust mapping reduces significantly such position error \iffalse from 0.16859 to -0.0434\fi to maintain the position. Consequently, the compensation scheme provides a better feed-forward term which leads to a smaller effort requested by the PD controller. Our model reduced the position errors along all directions when the most compact configuration of the Foldable Drone is assumed and its deployment provided a better trajectory tracking for the morphing platform.
The correct thrust mapping is now recovered, and a more stable and robust flight can be achieved.
\begin{table}[t!]
\begin{center}
\begin{tabular}{|C{1.5cm}|C{0.65cm}|C{0.65cm}||C{0.65cm}|C{0.65cm}||C{0.65cm}|C{0.65cm}|}
  \hline
  \multicolumn{7}{|c|}{\tabincell{c}{Experiments: Transitions in Forward Flight [m]}}\\
  \hline
 \multicolumn{1}{|c|}{\tabincell{c}{Morphology \\ Transitions}}  & $\mu_{\tiny X}$ & $\sigma_{\tiny X}$ & $\mu_{\tiny O_{\tiny n.c.}}$ & $\sigma_{\tiny O_{\tiny n.c.}}$ & $\mu_{\tiny O_{\tiny c.}}$ & $\sigma_{\tiny O_{\tiny c.}}$ \\
  \hline
  X - O$_{n.c.}$ & 0.0772 & 0.0414 & 0.2172 & 0.0787 & - & - \\
  \hline
  O$_{n.c.}$ - \textbf{O}$\mathbf{_{c.}}$ & - & - & 0.2339 & 0.050 & \textbf{0.1210} & 0.045 \\
  \hline
  X - \textbf{O}$\mathbf{_{c.}}$ & 0.0898 & 0.0452 & - & - & \textbf{0.1389} & 0.0482 \\
  \hline
  \hline
  X - O$_{n.c.}$ & 0.1211 & 0.0426 & 0.2690 & 0.0681 & - & - \\
  \hline
  O$_{n.c.}$ - \textbf{O}$\mathbf{_{c.}}$ & - & - & 0.2134 & 0.0452 & \textbf{0.1318} &  0.0611\\
  \hline
  X - \textbf{O}$\mathbf{_{c.}}$ & 0.0975 & 0.0363 & - & - &  \textbf{0.1038} & 0.0439 \\
  \hline
\end{tabular}
\caption{Statistics for the position error in forward flight for the canonical and most compact morphologies. Mean $\mu$ and standard deviation $\sigma$ for the euclidean distance in meters between the estimated position and the reference state. The first block considers flight in a circle with constant height, the second block concerns flight in a circle with varying height. \iffalse Note: ``c." stands for compensation and, ``n.c." stands for no compensation.\fi}
\label{ERR_pose_ff}% SOTA with Test
\end{center}
\vspace{-2mm}
\end{table}
\vspace{-2mm}
\section{Discussion}\label{disc}

%The role of morphing drones is always increasing for their versatile structure to perform specific tasks.
%Consequently, the need for thrust mapping of this kind will be increasingly requested, not only by different types of morphing quadrotors but also by drones that carry changing loads or a robotic arm.
%The presented compensation scheme for the specific platform of the Foldable Drone\cite{falanga2018foldable} can be extended to other multirotors whenever the analyzed geometric occlusions are present on the vehicle.
To tackle the problem of the loss of thrust due to the interaction between the propellers and the main body of a quadrotor, different approaches from the proposed one are possible.
For example, an alternative solution is the addition of an integral action on the already existing position controller, to compensate for deviations from its nominal hover position. 
However, such a solution has some drawbacks. 
The integral action can bring precision to slow systems, but when the imposed value varies rapidly and significantly, the integrator slows down the whole system, producing very often oscillations. 
Moreover, the integral action is effective only with a constant reference (in hovering), whereas, for the case of a time-varying set-point, this effectiveness is almost lost.
In addition to this limitation, the integrator also depends on the error of the state. Consequently, the integral action would start to have an impact only after the aerodynamic effect in analysis has affected the closed-loop system.
On the contrary, the proposed geometry-aware compensation scheme does not suffer from these issues. Our method exploits the actual physics of the phenomenon, without relying on any information of the state and without tuning any parameters. The action in feed-forward of our compensation scheme can overcome the loss of thrust immediately and the application of this compensation scheme only requires the knowledge of the geometry of the vehicle and the presence of the analyzed occlusions.
\vspace{-2mm}
\section{Conclusion}\label{sec:conc}
In this work, we presented a simple, yet effective analysis of the effects generated by partial occlusions between the propellers and the main body of a quadrotor. We leveraged this analysis to propose a novel geometry-aware compensation scheme. We proved its effectiveness by deploying it on a morphing quadrotor both in hovering and forward flight. Our model improved the position tracking in the configuration where its performance was affected the most (i.e., ``O" configuration), reaching almost the desired position and achieving more reliable, safe, and accurate flight.

Our work achieved a reduction of the position errors up to 20 cm in hovering, doubled the tracking accuracy of the drone in forward flight in the O configuration and, at the same time, reducing the effort of the PD controller, without additional power consumption and without increasing the execution time of the task that the drone is performing. 
\noindent
%Given the presence of the analyzed occlusions, our work can be extended to any multirotor aerial vehicle with either fixed or changing shape.

%\addtolength{\textheight}{-12cm}   % This command serves to balance the column lengths
                                  % on the last page of the document manually. It shortens
                                  % the textheight of the last page by a suitable amount.
                                  % This command does not take effect until the next page
                                  % so it should come on the page before the last. Make
                                  % sure that you do not shorten the textheight too much.

%%%%%%%%%%%%%%%%%%%%%%%%%%%%%%%%%%%%%%%%%%%%%%%%%%%%%%%%%%%%%%%%%%%%%%%%%%%%%%%%

%%%%%%%%%%%%%%%%%%%%%%%%%%%%%%%%%%%%%%%%%%%%%%%%%%%%%%%%%%%%%%%%%%%%%%%%%%%%%%%%

%%%%%%%%%%%%%%%%%%%%%%%%%%%%%%%%%%%%%%%%%%%%%%%%%%%%%%%%%%%%%%%%%%%%%%%%%%%%%%%%
%%%%%%%%%%%%%%%%%%%%%%%%%%%%%%%%%%%%%%%%%%%%%%%%%%%%%%%%%%%%%%%%%%%%%%%%%%%%%%%%%%%%%%
% For paper submission remove footnotesize
%%%%%%%%%%%%%%%%%%%%%%%%%%%%%%%%%%%%%%%%%%%%%%%%%%%%%%%%%%%%%%%%%%%%%%%%%%%%%%%%%%%%%%
{\footnotesize
\bibliographystyle{IEEEtran}
\bibliography{References}}
\end{document}